\documentclass[letterpaper, 10pt, conference]{ieeeconf}
\IEEEoverridecommandlockouts 
\usepackage{algorithm,algpseudocode}
\usepackage{balance}
\usepackage[utf8]{inputenc}
\usepackage[english]{babel}
\usepackage{pifont}
\usepackage{lscape}
\usepackage{longtable}
\usepackage{booktabs,tabularx}
\usepackage{graphicx}
\usepackage{multirow}
\usepackage{setspace}
\usepackage{hyperref}
\usepackage{amsmath,amssymb,amsfonts}
\usepackage{cuted}
\usepackage{cite}
\usepackage{color,soul,xcolor}
\usepackage[caption=false, font=footnotesize]{subfig}
\definecolor{myK}{RGB}{0, 0, 0}
\definecolor{myY}{RGB}{0, 0, 255}
\newenvironment{hlb}
{\color{myK}}
{}

\definecolor{aussiegold}{rgb}{0.9290, 0.6940, 0.1250}
\definecolor{aussiegreen}{rgb}{0.4660, 0.6740, 0.1880}
\definecolor{fireyred}{rgb}{0.8500, 0.3250, 0.0980}
\definecolor{airforceblue}{rgb}{0, 0.4470, 0.7410}
\definecolor{navygray}{rgb}{0.8,0.8,0.8}
\definecolor{newgray}{rgb}{0.5,0.5,0.5}
\definecolor{armygreen}{rgb}{0.4660,0.6740,0.1880}

\hypersetup{
    colorlinks=true,
    linkcolor=blue,
    filecolor=magenta,      
    urlcolor=blue,
    citecolor=magenta
}

\newtheorem{asm}{Assumption}

\title{Trust Modeling and Estimation in Human-Autonomy Interactions}
\author{Daniel A. Williams$^{1}$, Airlie Chapman$^{2}$, Daniel R. Little$^{3}$, Chris Manzie$^{1}$  
\thanks{$^{1}$\href{https://orcid.org/0000-0002-2983-7707}{Daniel A. Williams} and \href{https://orcid.org/0000-0002-9969-0982}{Chris Manzie} are with the Department of Electrical \& Electronic Engineering, The University of Melbourne, Australia. Williams is supported by a Commonwealth of Australia RTP Scholarship.}%
\thanks{$^{2}$\href{https://orcid.org/0000-0002-8946-552X}{Airlie Chapman} is with the Department of Mechanical Engineering,
        The University of Melbourne, Australia.}%
\thanks{$^{3}$\href{https://orcid.org/0000-0003-3607-5525}{Daniel R. Little} is with the Melbourne School of Psychological Sciences, The University of Melbourne, Australia.}
}

\begin{document}

\maketitle
\begin{abstract}
Advances in the control of autonomous systems have accompanied an expansion in the potential applications for autonomous robotic systems. 
The success of applications involving humans depends on the quality of interaction between the autonomous system and the human supervisor, which is particularly affected by the degree of trust that the supervisor places in the autonomous system.
Absent from the literature are models of supervisor trust dynamics that can accommodate asymmetric responses to autonomous system performance and the intermittent nature of supervisor-autonomous system communication.
This paper focuses on formulating an estimated model of supervisor trust that incorporates both of these features by employing a switched linear system structure with event-triggered sampling of the model input and output.
Trust response data collected in a user study with 51 participants were used identify parameters for a switched linear model-based observer of supervisor trust.
This yielded models corresponding to individuals, clusters of similar individuals, and the population.
The proposed model with cluster-based parameters may be suitable for augmenting communication interfaces for human-autonomous system interactions, allowing a supervisor's trust to be monitored with minimal self-reporting.
\end{abstract}

\section{Introduction}

\begin{figure*}[t]
    \centering
    \includegraphics[width=1.8\columnwidth]{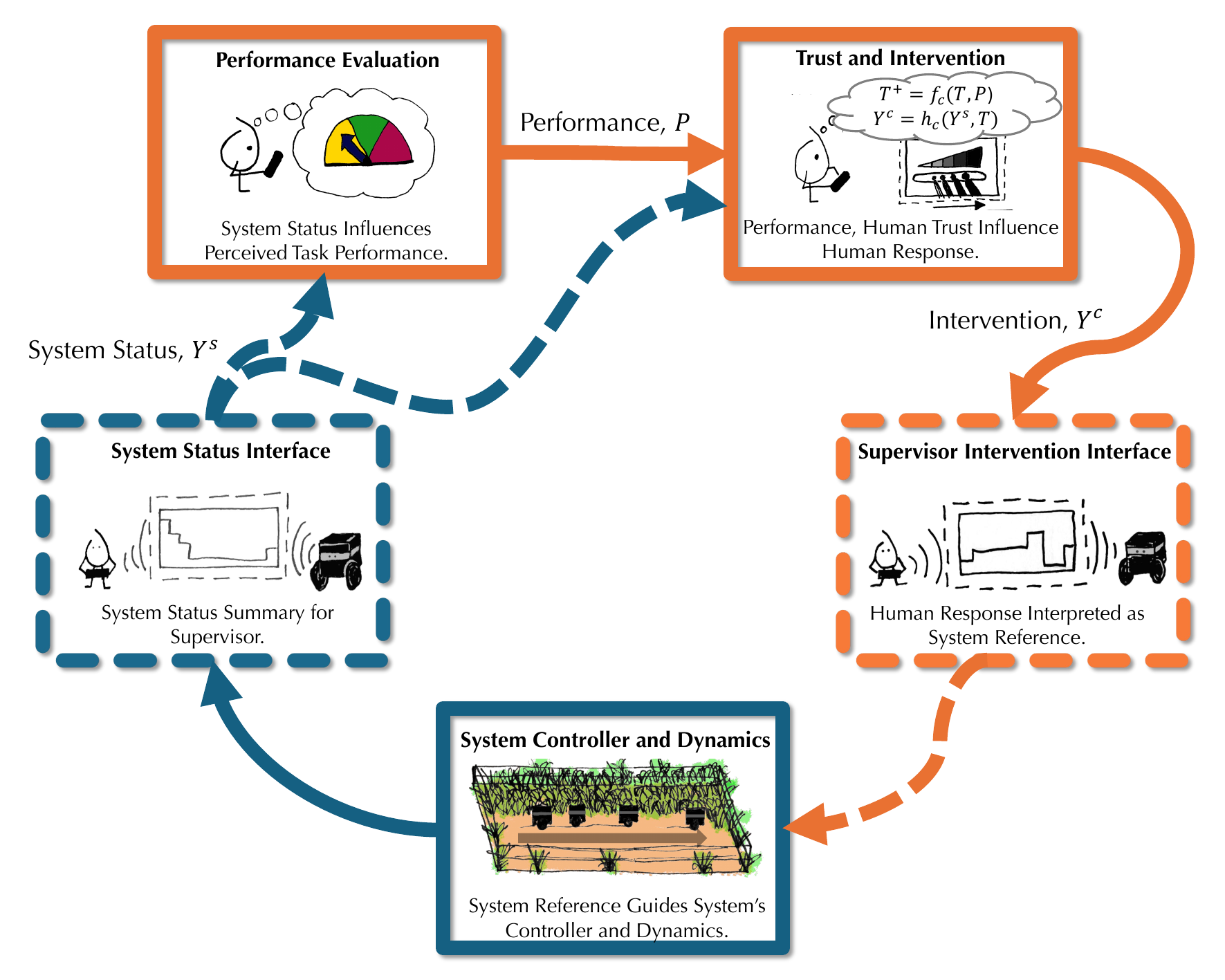}
    \caption{Overview of the trust framework.}
    \label{fig:framework}
\end{figure*}

With the ongoing development of autonomous systems, significant attention has focused on the deployment of autonomous robotic systems within human-machine teams \cite{drew_multi-agent_2021}.
These applications often involve human-on-the-loop decision making, in which a human supervisor delegates responsibility for a task to the autonomous system \cite{gebru_review_2022}.
In this way, the supervisor balances the cognitive load between the supervision of the autonomous system and other tasks \cite{kolling_human_2016}.
A notion of human trust in the autonomous system can describe the extent of the supervisor's willingness to delegate responsibility to the autonomous system \cite{hussein_reliability_2020}.
This can be influenced by the complexity of the task, environmental conditions, and the autonomous system's composition \cite{lewis_deep_2020}.
In turn, the supervisor's level of trust can affect the likelihood of supervisor intervention in the short term, and of reliance on the autonomous system in the long term \cite{wang_co-design_2017,wang_human_2023}.

To estimate human trust using systems theoretical techniques, a dynamic model for trust must first be specified \cite{williams_hri_2025}.
A common approach in trust modeling supposes a probabilistic relationship between the current task performance of the autonomous system and human trust, whether for an individual human \cite{mahani_bayesian_2020,nam_predicting_2017,chen_trust_2018,soh_multi_2020}, a cluster of individuals \cite{mcmahon_modeling_2020}, or a population \cite{xu_optimo_2015,chen_trust_2019}.
These models can accommodate uncertainties inherent in the definition and measurement of human trust, however their structures can obscure the relationships between model variables and reduce model interpretability \cite{rudin_interpretable_2022}.

Alternatively, deterministic trust models offer greater transparency about variable relationships and lend themselves to the design of simpler estimators of trust for human-machine interfaces at the cost of reducing their verisimilitude.
Such models have been developed for clusters of individuals \cite{akash_dynamic_2017,liu_clustering_2021} or a population \cite{yang_evaluating_2017} that use linear dynamics to describe relationship between the input (task performance) and state (human trust).
An important feature of these models is that they apply the same parameter values across the entire range of possible input values, hence trust responds symmetrically to positive and negative values of task performance.
Consequently, linear trust models may be unable to describe the trust responses of individuals who exhibit \textit{non-symmetric} trust dynamics (e.g. when trust is `quick to lose, slow to gain' \cite{robinette_effect_2017,wang_survey_2024}).
An open question remains: are there benefits to incorporating some degree of non-linearity to improve trust predictions?

After specifying a given model structure for trust, identifying appropriate model parameter values requires measurements of trust during interactions.
Motivated by existing results about trust in social psychology, a considerable body of literature has emerged around measuring trust in autonomous systems \cite{shahrdar_survey_2019}.
Several studies gauge trust explicitly through self-reporting \cite{lee_trust_1992,jian_foundations_2010}, either as an absolute value \cite{schaefer_measuring_2016} or a relative change \cite{desai_impact_2013}.
Nonetheless, self-reporting methods can disrupt interaction if they are frequent or cognitively taxing.
For many applications it may not be practical to continuously collect data via self-reporting to identify an individual's trust model parameters, particularly if these are influenced by unmodeled phenomena (e.g. an individual's emotional state).

If individuals can be grouped so that their trust responses are characterized by common model parameters, it may be easier to identify the group rather than the set of model parameters.

    In this approach, a clustering algorithm \cite{lloyd_kmeans_1982,navarro2006modeling,little2009beyond} can be used to generate \textit{clusters} varying in size from a single individual to a whole population.
These clusters can then enable ready identification of a sufficiently accurate model of trust \cite{liu_clustering_2021,vella_individual_2022}.

Having access to such a model could allow autonomous systems to develop trust estimates and incorporate these into interactions, potentially reducing perceived barriers to more effective cooperation \cite{matthews_individual_2020}.

Another important attribute of supervisor-autonomous system interaction is the intermittent nature of communication between the two parties \cite{kolling_human_2016}.
To reduce costs arising from continuous transmission over communication channels, the autonomous system may only send an update on the task's completion status when there is an noticeable difference from the last update.
Similarly, to avoid micro-managing the supervisor may decide to only issue a new intervention when it would improve autonomous system performance \cite{joo_formalizing_2019}.
Such intermittency suggests that the supervisor-autonomous system communication interfaces can be represented as hybrid systems with event-triggered samples.

In event-triggered sampling, a new update is triggered only when the  sampler's error exceeds a pre-determined threshold (termed an `event') \cite{williams_asynchronous_2024}.
This method of sampling can replace periodic surveys of a human participant's trust, reducing communication transmissions \cite{williams_tac_nodate}.
In \cite{williams_hri_2025} a mathematical framework is proposed that represents trust-driven interactions between a human and autonomous system as a closed feedback loop with event-triggered communication, however the identification of clusters of individuals and their trust model from collected data remains unexplored.

In light of these gaps surrounding the modeling of trust in human-autonomy interactions, this paper makes the following contributions:
\begin{enumerate}
    \item We propose a switched linear model structure to represent a supervisor's potentially \textit{non-symmetric} trust response that accommodates state saturation and event-triggered sampling of model input and output signals.
    \item We use the switched linear model structure to investigate whether individuals can be effectively grouped into \textit{clusters} with similar trust characteristics.
    \item We use data collected from a user study with 51 participants to identify trust model parameters for individuals, clusters and a population, and compare the performance of all three types of models.
\end{enumerate}

\section{Trust Modeling}\label{sec:trustmodeling}

We will ground the discussion of trust by considering an interaction between a human supervisor and an autonomous system.
As depicted in Figure \ref{fig:framework}, the interaction can be represented as a closed loop interconnection of five subsystems according to the framework proposed in \cite{williams_tac_nodate}.
Two subsystems describe the human supervisor's involvement (Performance Evaluation, and Trust and Intervention Dynamics).
One subsystem represents a supervisor-to-autonomous system communication interface (Supervisor Intervention Interface) while another represents an autonomous system-to-supervisor communication interface (System Status Interface).
The final subsystem captures the autonomous system's controller and dynamics (System Controller and Dynamics)
In this paper we focus on the `Trust and Intervention' subsystem, and we begin by considering the design and selection of candidate models using a class of non-linear systems.

\subsection{Model Structure}

Motivated by the body of work linking the performance of autonomous systems to supervisor trust \cite{lee_trust_1992,akash_dynamic_2017,liu_clustering_2021}, we propose nonlinear dynamics for supervisor trust $\mathbf{T}$ driven by the autonomous system's performance $\mathbf{P}$ and an exogenous environmental input $w$.

We assume that $\mathbf{T}$ evolves in the closed domain $\tilde{C}_T:=[\mathbf{T_{min}},\mathbf{T_{max}}]\subset \mathbb{R}$, which without loss of generality can be normalized to the range $[0\%,100\%]$, that $\mathbf{P}$ exists in the closed domain $\tilde{C}_P:=[\mathbf{P_{min}},\mathbf{P_{max}}]\subset \mathbb{R}$, and that $w\in\tilde{C}_w\subset\mathbb{R}$ remains constant for a given interaction.
Note that these variables can be generalized to multi-dimensional quantities to accommodate vector-valued metrics.
We next define a map $f_c:\tilde{C}_P \times \tilde{C}_T^{n_T}\times \tilde{C}_w \rightarrow \tilde{C}_T$ such that trust is updated in discrete time by $\mathbf{T}[k+1] = f_c(\mathbf{P}[k],\{\mathbf{T}[k+1-j]\}_{j=1}^{n_T},w)$, where $n_T\in\mathbb{Z}$ is the size of the model's memory element. 
The specific choice of $f_c$ is influenced by a supervisor's individual experiences, background and personality \cite{matthews_individual_2020}.

As a candidate for $f_c$, we propose a switched linear system model structure. 
Switched linear systems can represent a system's dynamics as one of several modes governed by a linear equation, with mode switching controlled by a switching signal.
The benefit of using such a model for representing trust is that one can choose the linear equations to reproduce an asymmetric response of trust to system performance, while retaining a sense of interpretability that is obscured with other non-linear systems.
To this end, consider a population of $n_s\in\mathbb{N}$ supervisors. For the $i$th supervisor, $i\in\{1,...,n_s\}$, we define the trust update
\medmuskip=-1mu
\thinmuskip=-2mu
\thickmuskip=-2mu
\nulldelimiterspace=-1pt
\scriptspace=0pt
\begin{align}
    \mathbf{T}[k+1] &= \sum_{j=1}^{n_T} A_{j,\sigma[k],i}\mathbf{T}[k+1-j] + B_{\sigma[k],i}\mathbf{P}[k]+G_{\sigma[k],i}w,\label{eq:trustupdate_sls}
\end{align}
\thinmuskip=3mu
\medmuskip=4mu
\thickmuskip=5mu
\nulldelimiterspace=0.5pt
\scriptspace=0.5pt
where $\{\{A_{j,\sigma,i}\}_{\sigma=1}^{n_\sigma}\}_{j=1}^{n_T}\in\mathbb{R}$, $\{B_{\sigma,i}\}_{\sigma=1}^{n_\sigma}\in\mathbb{R}$, and $\{G_{\sigma,i}\}_{\sigma=1}^{n_\sigma}\in\mathbb{R}$, are supervisor-specific model coefficients, and $\sigma[k]\in\{1,...,n_{\sigma}\}$, ${n_\sigma}\in\mathbb{N}$, denotes the system's mode at time step $k$.
For convenience, we will use the notation $A_{m,i}=A_{j,m,i}$ when $n_T=1$, and $\bar{A}_{m}=A_{j,m,i}$ when $n_T=1$ and $n_C = 1$.

The resulting intervention $\mathbf{Y}^\mathbf{c}\in \tilde{C}_o\subset\mathbb{R}^{n_o}$ by the supervisor is given by
\begin{align}
    \mathbf{Y}^\mathbf{c}[k] &= C_{\sigma[k],i}\mathbf{T}[k] + H_{\sigma[k],i}w\label{eq:trustoutput_sls}
\end{align}
where the subsystem matrices $\{C_{\sigma,i}\in\mathbb{R}^{n_o}\}_{\sigma=1}^{n_\sigma}$ and $\{H_{\sigma,i}\in\mathbb{R}\}_{\sigma=1}^{n_\sigma}$ are also selected using $\sigma[k]$.

\subsection{Model Implementation}

\begin{figure}[t]
    \centering
    \includegraphics[width=\columnwidth]{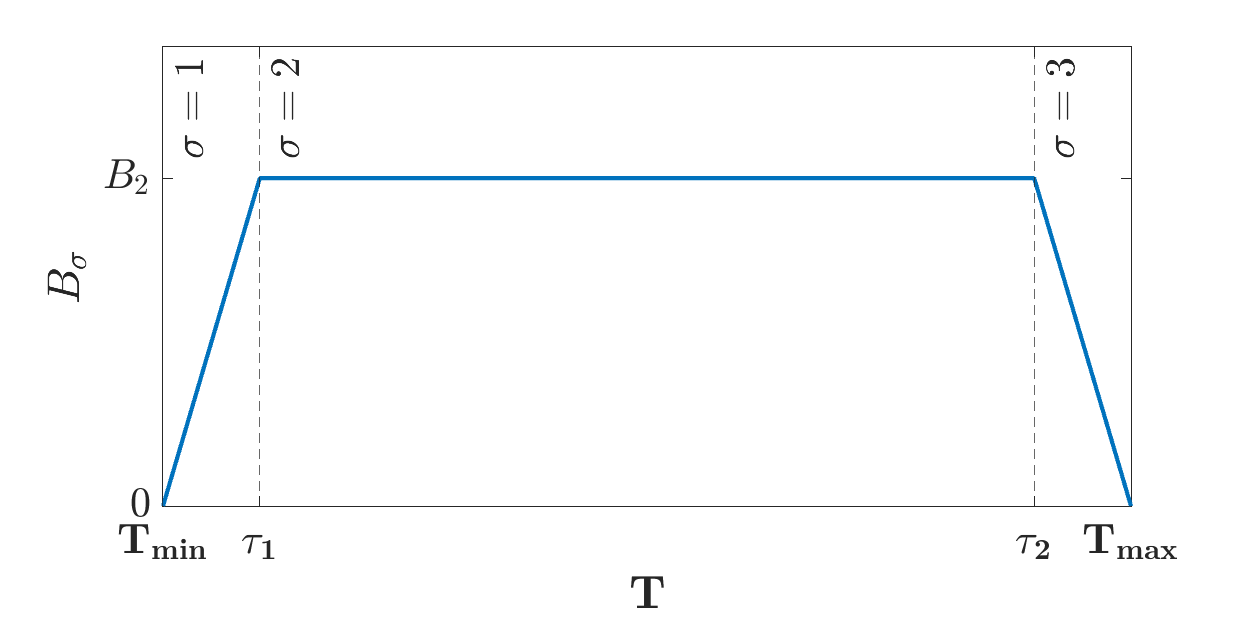}
    \caption{The coefficients of autonomous system performance for $\sigma\in\{1,2,3\}$; note that the graph is identical for $\sigma\in\{4,5,6\}$ with $B_5$ substituted for $B_2$.}
    \label{fig:Bplots}
\end{figure}

To ensure that each mode has sufficient data for training the corresponding parameters, we specify $n_\sigma = 6$ with the mode-switching signal $\sigma[k]$ given by
\begin{align} 
    \sigma[k] &= \begin{cases}
        1,& \mathbf{P}[k]\in[\mathbf{P_{min}},\mathbf{P^*}),\mathbf{T}[k]\in(\mathbf{\tau_2},\mathbf{T_{max}}],\\
        2,& \mathbf{P}[k]\in[\mathbf{P_{min}},\mathbf{P^*}),\mathbf{T}[k]\in[\mathbf{\tau_1},\mathbf{\tau_2}],\\
        3,& \mathbf{P}[k]\in[\mathbf{P_{min}},\mathbf{P^*}),\mathbf{T}[k]\in[\mathbf{T_{min}},\mathbf{\tau_1}),\\
        4,& \mathbf{P}[k]\in[\mathbf{P^*},\mathbf{P_{max}}],\mathbf{T}[k]\in(\mathbf{\tau_2},\mathbf{T_{max}}],\\
        5,& \mathbf{P}[k]\in[\mathbf{P^*},\mathbf{P_{max}}],\mathbf{T}[k]\in[\mathbf{\tau_1},\mathbf{\tau_2}],\\
        6,& \mathbf{P}[k]\in[\mathbf{P^*},\mathbf{P_{max}}],\mathbf{T}[k]\in[\mathbf{T_{min}},\mathbf{\tau_1}).
    \end{cases}  
\label{eq:switcher}
\end{align}
\begin{figure}[t]
    \centering
    \includegraphics[width=\columnwidth]{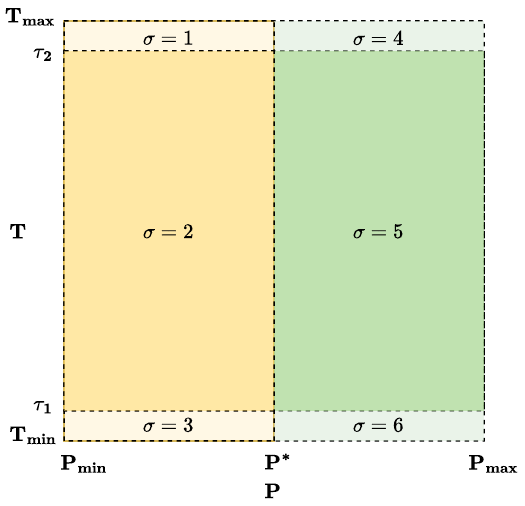}
    \caption{The proposed six modes of trust.}
    \label{fig:sixmodes}
\end{figure}
\begin{table}[t]
\centering
\begin{tabular}{|c|c|cc|c|c|}
\hline
\multirow{2}{*}{$\sigma$} & $n_T=1$ & \multicolumn{2}{c|}{$n_T=2$} & \multirow{2}{*}{$B_{\sigma,i}$} & \multirow{2}{*}{$G_{\sigma,i}$}\\ \cline{2-4}
 & $A_{\sigma,i}$ & \multicolumn{1}{c|}{$A_{1,\sigma,i}$} & $A_{2,\sigma,i}$ &  &\\ \hline
1 &$1-\epsilon$  & \multicolumn{1}{c|}{$1-\epsilon$} &$0$  & $\gamma\left(1- \frac{\mathbf{T}-\mathbf{\tau_2}}{\mathbf{T_{max}}-\mathbf{\tau_2}}\right)$ & $0$\\ \hline
2 &$\alpha$  & \multicolumn{1}{c|}{$\alpha_1$} &$\alpha_2$  &$\gamma$  &$\kappa$ \\ \hline
3 &$1+\epsilon$  & \multicolumn{1}{c|}{$1+\epsilon$} &$0$  &$\gamma\left(1-\frac{\mathbf{\tau_1}-\mathbf{T}}{\mathbf{\tau_1}-\mathbf{T_{min}}}\right)$  & $0$\\ \hline
4 &$1-\epsilon$  & \multicolumn{1}{c|}{$1-\epsilon$} &$0$  &$\delta\left(1-\frac{\mathbf{T}-\mathbf{\tau_2}}{\mathbf{T_{max}}-\mathbf{\tau_2}}\right)$  &$0$\\ \hline
5 &$\beta$  & \multicolumn{1}{c|}{$\beta_1$} &$\beta_2$  &$\delta$  &$q$\\ \hline
6 &$1+\epsilon$  & \multicolumn{1}{c|}{$1+\epsilon$} &0  &$\delta\left(1-\frac{\mathbf{\tau_1}-\mathbf{T}}{\mathbf{\tau_1}-\mathbf{T_{min}}}\right)$  & $0$\\ \hline
\end{tabular}
\caption{Individual state-space model parameters for $n_T\in\{1,2\}$, where $\alpha,\alpha_1,\alpha_2,\beta,\beta_1,\beta_2,\gamma,\delta,\kappa,q$ are identified from data such that the model is stable for $\sigma\in\{2,5\}$, and $\epsilon, \tau_1, \tau_2$ are specified.}\label{tab:ind}
\end{table}
As illustrated in Figure \ref{fig:sixmodes}, the parameters $\mathbf{P^*}\in\mathbb{R}$, $\mathbf{\tau_1},\mathbf{\tau_2}\in\mathbb{R}$ serve to partition $\{\tilde{C}_T \times \tilde{C}_P\}$ as follows.
We first divide $\{\tilde{C}_T \times \tilde{C}_P\}$ into two regions according to the polarity of $\mathbf{P}-\mathbf{P^*}$ (either negative or non-negative when $\mathbf{P}$ is scalar). 
This permits the trust model to respond differently when $\mathbf{P}$ is greater than $\mathbf{P^*}$ or less than $\mathbf{P^*}$.
We then define $\mathbf{\tau_1}$ and $\mathbf{\tau_2}$ as soft boundaries for $\mathbf{T}$ to ensure that $\mathbf{T}$ does not continue increasing above $\mathbf{T_{max}}$ (or decreasing below $\mathbf{T_{min}}$).
This creates two sub-regions within $\mathbf{T}\in[\mathbf{\tau_1},\mathbf{\tau_2}]$ denoted as modes 2 and 5, and four sub-regions outside these boundaries (modes 1, 3, 4, and 6).
To ensure that $\mathbf{T}$ remains within $[\mathbf{T_{min}},\mathbf{T_{max}}]$ in the latter modes, we choose the state-space model coefficients as per Table \ref{tab:ind}.
To this end we seek some $\alpha,\beta\in\mathbb{R}$ if $n_T=1$ (or $\alpha_1,\alpha_2,\beta_1,\beta_2\in\mathbb{R}$ if $n_T=2$), $\gamma,\delta,\kappa,q\in\mathbb{R}$, and $\epsilon\in(0,1)$.
For $\sigma\in\{1,3,4,6\}$, the choice of $A_\sigma$ drives $\mathbf{T}$ back into the range $[\mathbf{\tau_1},\mathbf{\tau_2}]$. 
At the same time, the value of $B_\sigma$ is tapered off to reduce the influence of performance as $\mathbf{T}$ approaches $\mathbf{T_{max}}$ from below (or $\mathbf{T_{min}}$ from above) as depicted in Figure \ref{fig:Bplots}.

\begin{figure*}[t]
    \centering
    \includegraphics[width=1.7\columnwidth]{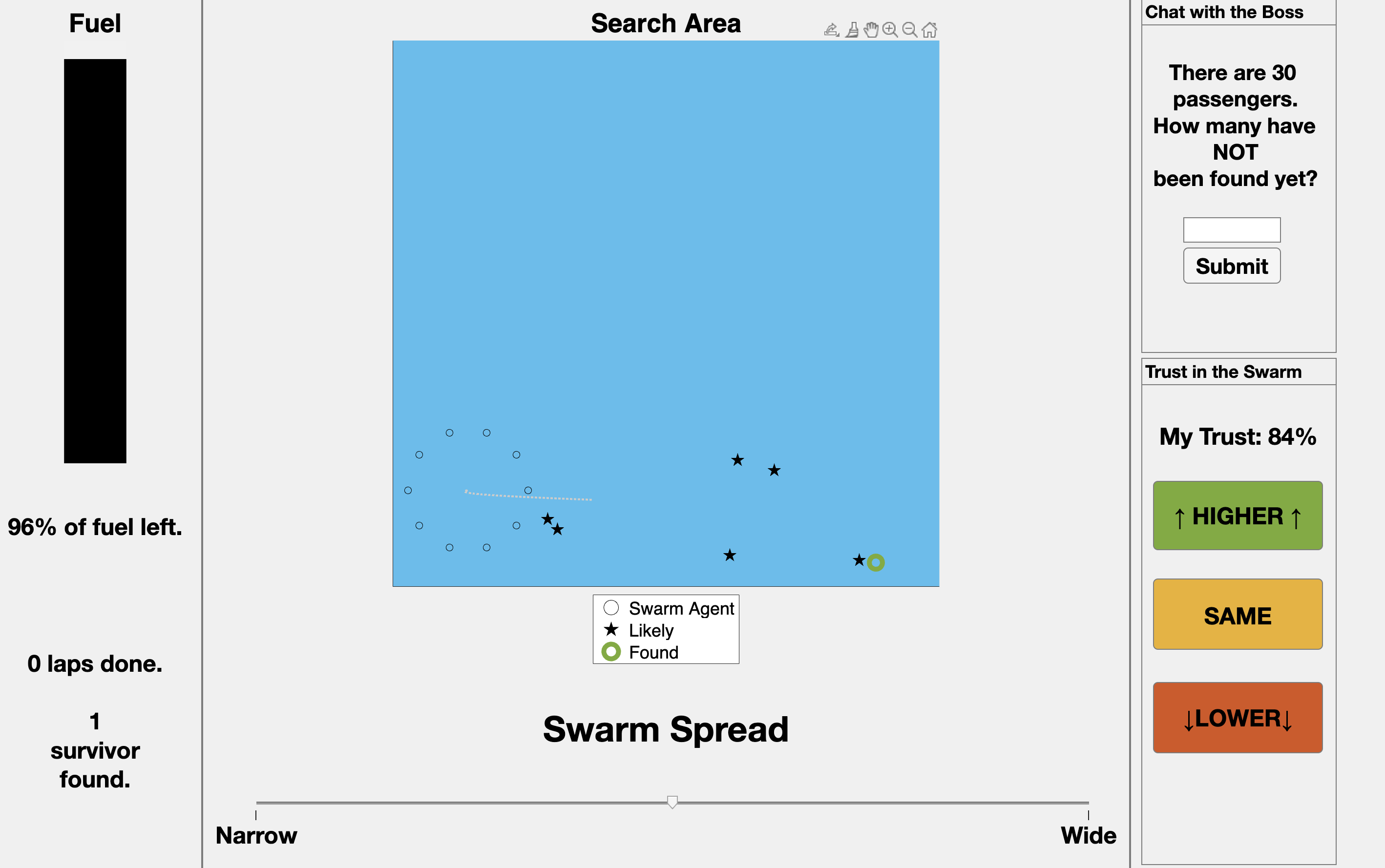}
    \caption{The user study's simulation interface.}
    \label{fig:simint}
\end{figure*}

\subsection{Parameter Identification \label{sec:paramid}}

Let there be $n_s$ supervisors divided among $n_c\leq{n_s}$ groups.
If $n_c=n_s$ then an individual model of supervisor trust is learned.
If $n_c=1$ a population-wide model is learned.
If $n_c\in(1,n_s)$ then a cluster-based model is learned.
Let the signals $\mathbf{P}_{m,i}, \mathbf{T}_{m,i},\mathbf{Y}^{\mathbf{c}}_{m,i}$, $m\in\{1,...,6\}$, represent the intervals of $\mathbf{P}$, $\mathbf{T}$, and $\mathbf{Y^c}$ partitioned in time according to the switching mode signal and concatenated for all supervisors belonging to the $i$th group. 
The following assumption guarantees persistency of excitation for the collected data.
\begin{asm}\label{asm:suff}
For all $i\in\{1,...,n_c\}$ and $m\in\{2,5\}$, the signals $\mathbf{P}_{m,i}$, $\mathbf{T}_{m,i}$, and $w$ satisfy $\mathrm{det} (M^\prime M)\neq 0$, where $M=\begin{bmatrix}
    \mathbf{P}_{m,i} &\mathbf{T}_{m,i} & w
\end{bmatrix}$.
\end{asm}
For $i\in\{1,...,n_c\}$, define the variables $\mathbf{\Theta}_i=(\{\alpha_{j,i}\}_{j=1}^{n_T},\gamma_i,\kappa_i)$, 
$\mathbf{\Phi}_i=(\{\beta_{j,i}\}_{j=1}^{n_T},\delta_i,q_i)$, 
and $\mathbf{\Psi}_i = (\{C_{m,i}\}_{m=1}^{n_\sigma},\{H_{m,i}\}_{m=1}^{n_\sigma})$, and the objective functions
\begin{align}
J_1(\mathbf{\Theta}_i) =&\sum_{k:\sigma[k]=2}(\mathbf{T}_{2,i}[k+1]  - \sum_{j=1}^{n_T}\alpha_{j,i}\mathbf{T}_{2,i}[k+1-j] \nonumber\\&- \gamma_i\mathbf{P}_{2,i}[k]-\kappa_iw)^2,\label{eq:j1}\\
J_2(\mathbf{\Phi}_i)=&\sum_{k:\sigma[k]=5}(\mathbf{T}_{5,i}[k+1]  - \sum_{j=1}^{n_T}\beta_{j,i}\mathbf{T}_{5,i}[k+1-j] \nonumber\\&- \delta_i\mathbf{P}_{5,i}[k]-q_iw)^2,\label{eq:j2}\\
J_3(\mathbf{\Psi}_i)=&\sum_{m=1}^{n_\sigma}\sum_{k:\sigma[k]=m}(\mathbf{Y}^\mathbf{c}_{m,i}[k+1]  - C_{m,i}\mathbf{T}_{m,i}[k]\nonumber\\&-H_{m,i}w)^2.\label{eq:j3}
\end{align}
The parametrization of \eqref{eq:trustupdate_sls}--\eqref{eq:trustoutput_sls} for the $i$th group can be found by solving the three optimization problems
\begin{align}
    \mathbf{\Theta}^*_i =&\,\mathrm{arg}\min_{\mathbf{\Theta}_i}  J_1(\mathbf{\Theta}_i),\label{eq:optimAlpha}\\
    &\mathrm{s.t.}\,|\lambda_{m,i}|\leq1,\,m\in\{1,...,n_\sigma\},\label{eq:Alpha_constraint}\\
    \mathbf{\Phi}^*_i =&\,\mathrm{arg}\min_{\mathbf{\Phi}_i}J_2(\mathbf{\Phi}_i),\label{eq:optimBeta}\\
    &\mathrm{s.t.}\,|\mu_{m,i}|\leq1,\,m\in\{1,...,n_\sigma\},\label{eq:Beta_constraint}\\
    \mathbf{\Psi}^*_i =& \,\mathrm{arg}\min_{\mathbf{\Psi}_i}J_3(\mathbf{\Psi}_i),\label{eq:optimC}
\end{align}
where $\lambda_{m,i}$ is the $m$th root of the polynomial equation $\lambda-\sum_{j=1}^{n_T}\alpha_{j,i}\lambda^{1-j}=0$, and $\mu_{j,i}$ is the $j$th root of the polynomial equation $\mu-\sum_{j=1}^{n_T}\beta_{j,i}\mu^{1-j}=0$.
To ensure that the identified trust model is stable in modes 2 and 5, we impose the constraints \eqref{eq:Alpha_constraint} and \eqref{eq:Beta_constraint}.
These ensure that the $\mathbf{P}$-to-$\mathbf{T}$ transfer function for \eqref{eq:trustupdate_sls} has poles of magnitude less than or equal to 1.
Note that in \eqref{eq:j1}--\eqref{eq:j3} an equal weighting is given to the data contributing to $\mathbf{P}_{m,i}$, $\mathbf{T}_{m,i}$, and $\mathbf{Y}^{\mathbf{c}}_{m,i}$, however this does not imply that the individuals have an equal-sized influence on the model parameter values.

\subsection{Towards Clustered Trust Responses}\label{sn:clustertheory}
It is potentially advantageous to define a set of $n_c\in(1,n_s)$ trust models that represent common trust dynamics within a population corresponding to distinct clusters of individuals.
This can be achieved by identifying parameters for individuals' trust responses, defining an embedding space to represent these individuals, grouping individuals into clusters using their embeddings, aggregating data for each cluster, and identifying the clusters' parameters.

After identifying individuals' trust model parameters in Section \ref{sec:paramid}, we construct a vector to represent each individual in an embedding space.
For a given mode $\sigma$, the $i$th individual's trust response can be characterized by the transfer functions $\frac{B_{\sigma,i}}{s-A_{\sigma,i}}$ and $\frac{G_{\sigma,i}}{s-A_{\sigma,i}}$.
\begin{hlb}The poles at $s=A_{\sigma,i}$ affect the dynamic response, while the numerators determine the static gains.
As the individual's trust response is expected to remain in modes 2 and 5 for most of the session duration,\end{hlb} we thus define for the $i$th individual the embedding vector $v_i = \begin{bmatrix} \alpha^* &\beta^* \end{bmatrix}\in\mathbb{R}^{2n_T}$, with $\alpha^*=\begin{bmatrix}
    \alpha_{j}
\end{bmatrix}_{j=1}^{n_T}\in\mathbb{R}^{n_T}$
and $\beta^*=\begin{bmatrix}
    \beta_{j}
\end{bmatrix}_{j=1}^{n_T}\in\mathbb{R}^{n_T}$. 

We next use the $k$-means clustering algorithm \cite{lloyd_kmeans_1982} to find clusters of individuals in the embedding space.
We perform the clustering for a range of values of $k$ and record the total sum of distances from \begin{hlb}each vector $v_i$\end{hlb} to its nearest cluster centroid.
As the algorithm can converge to multiple local optima, we repeat the algorithm using the same $k$ value from multiple initial conditions and retain the solution yielding the lowest total sum of distances for each $k$ value.
We then use the lowest total sums of distances to identify the smallest value of $k$ that achieves suitably low within-cluster distances while avoiding the creation of singleton clusters for outliers.

After selecting an appropriate \begin{hlb}$k$ value, we determine the cluster centroid using a weighted mean of the parameter values\end{hlb} for individuals within the cluster.
By associating an individual with an existing cluster, we can use the cluster centroid to estimate the trust response for that individual.

\section{User Study}\label{sec:userstudy}
A common test for human-autonomous system interaction is the foraging task \cite{nam_predicting_2017,walker_neglect_2012}.
We propose a variant in which an autonomous system of robotic agents searches for 30 survivors uniformly distributed in a square region of side length $50$ km.
Inspired by the approach of \cite{breslow_dynamic_2014}, the study was conducted via the simulation interface depicted in Figure \ref{fig:simint}.
The autonomous system's agents move in a ring-shaped formation, with the centroid tracking a sinusoidal trajectory.

The time available for the formation to search the region is constrained by the quantity of fuel carried by the agents, which has a constant rate of depletion.
The formation can complete more than one lap of the region if there is fuel remaining.
The area within the region that can be inspected by the formation is determined by the speed of the formation's centroid.
This speed is varied proportionally to the formation radius, which is chosen by the supervisor using a slider scale at the bottom of the interface.
The slider scale permits formation radii between $1$ km (``Narrow'') and $10$ km (``Wide''), with a default radius of $5.5$ km.

The mechanism by which the formation detects a survivor is determined by the distances from the survivor to each agent and is detailed in \cite{williams_tac_nodate}.
When more than one agent is within $2$ km of a survivor's position, the formation's confidence in having detected that survivor increases, i.e. a smaller formation radius promotes better survivor detection.
This invokes a trade-off between the speed of search and the likelihood of confirmed detections of survivors.
The supervisor must intermittently assess environmental conditions and known positions of survivors in order to select a suitable formation radius.
Supervisor performance is measured as a weighted sum of the number of survivors found by the formation and a supplementary score on a secondary task.
This metric is made available at the conclusion of each session to allow supervisors to compare their performance over the series of missions.

Fifty-one participants were recruited without reimbursement for participation.
Ethical approval for the study was granted by the Office of Research Ethics and Integrity at the authors' university with reference 2023-27715-45206-3.

\subsection{Method}
Each participant undertook a series of sessions (a one-minute practice followed by two full sessions) using the interface, which showed the autonomous system agents' positions, the formation centroid's recent trajectory, and suspected and confirmed survivor positions.
After every session, participants rested for 30 seconds.

Before each session participants completed the abridged 14-question TPSHRI scale \cite{schaefer_measuring_2016}, a standard method of measuring absolute trust in human-robot interaction informed by psychological studies.
The TPSHRI scale responses for each participant were used to measure absolute trust values, with the initial trust value calculated as the mean of TPSHRI scores not set as `N/A', and set to $50\%$ by default if users set all scores as `N/A'.
During each session, participants were instructed to attend to three tasks:
\begin{enumerate}
    \item \textbf{Supervise the autonomous system's search and intervene by changing the formation radius (``Swarm Spread") $\mathbf{Y^{c}}$ } (cf. \cite{nam_predicting_2017}). 
    The supervisor intervention was sampled as per \cite{williams_asynchronous_2024}, such that if a supervisor's change in the radius was above a threshold, a new sampling event was triggered.
    Changes to the radius were subject to a minimum waiting period to avoid Zeno-type behavior.
    The parameters of the sampler subsystem were set to $e_u[k]:=\mathbf{\hat{Y}^c}[k]-\mathbf{Y^c}[k]$, $V_c(\mathbf{Y^c},e_u):=(\mathbf{Y^c})^2+\frac{e_u^2}{100}$, $W_u(e_u) = 10^3e_u^2$, and $\tau_c=0.5$.\\
    \item \textbf{Self-report any changes in trust in the autonomous system} by registering an increase in trust, no change in trust, or a decrease in trust respectively (cf. \cite{desai_impact_2013}).
    Trust was adjusted in $\pm$5\% increments within the range $[\mathbf{T_{min}}, \mathbf{T_{max}}]$, with these samples used to reconstruct a signal for $\mathbf{T}[k]$.  
    Motivated by the trust surveying approach of \cite{lu_eye_2019}, the simulation was paused temporarily if 45 s had passed since the last trust report (thus enforcing a minimum sampling frequency).\\
    \item \textbf{Complete simple two-digit subtractions \cite{crandall_validating_2005} presented using a chat box \cite{dahiya_interruptions_2023}}. Points were awarded for correct answers to motivate continued engagement with the interface during periods of lower cognitive burden. 
    This task could be completed alongside the primary task at the supervisor's discretion, with a minimum waiting period of 10 s between responses enforced to avoid the supervisor neglecting the primary task.
\end{enumerate} 
Concurrent with the collection of data from the participant, the system status signal $\mathbf{Y^s}[k]$ was recorded as the percentage of the 30 survivors that were found.
This status signal was sampled using the event-triggered sampler in \cite{williams_asynchronous_2024}, with $e_m[k]:=\mathbf{\hat{Y}^s}[k]-\mathbf{Y^s}[k]$, $V_p(\mathbf{Y^s},e_m):=(\mathbf{Y^s})^2+\frac{e_m^2}{100}$, $W_p(e_m) = 10^3e_m^2$, and $\tau_p=0.5$.
For each individual, the data from the first full session were aggregated into a training set, while the data from the second full session were allocated to a test set. 
The performance metric was calculated as 
\begin{align}
    \mathbf{P}[k] &= r_s[k] - r_l[k],
\end{align} where $r_s[k] = \frac{\mathbf{\hat{Y}^s}[k] - \mathbf{\hat{Y}^s}[k-n_q]}{n_q}$ denotes the short-term rate of survivors found over a recent memory window of length $n_q>0$ and $r_l[k] = \frac{\mathbf{\hat{Y}^s}[k]}{2k}$ the long-term average rate of finding survivors.
This definition of performance is compatible with the structure of attributive emotion types in \cite{ortony2022cognitive}, under which a supervisor's impression of system performance results from the supervisor focusing on recent performance (captured by $r_s[k]$) relative to expectations approximated by $r_l[k]$.
The memory length $n_q$ was treated as a tunable hyperparameter.
The remaining variables were set to $w=1$, $\mathbf{P^*}=0$, $\mathbf{T_{min}}=0\%$, $\tau_1=0.1\%$, $\tau_2=99.9\%$, $\mathbf{T_{max}}=100\%$, and $\epsilon=10^{-2}$.

\subsection{Results}

\begin{algorithmic}
\begin{algorithm}
\caption{Find model parameters $(n_q^*,\mathbf{\Theta}^*,\mathbf{\Phi}^*,\mathbf{\Psi}^*)$.\label{alg:findnq}}
\State $E^* \gets 10^{10}\cdot\mathbf{1}_{n_c\times 1}$
\For {cluster $i\in\{1,...,n_c\}$}

    \For{$n_q\in\{5,10,15,20,30,45,60,75,90,120\}$}
        \State $(\mathbf{P}_{m,i},\mathbf{T}_{m,i},\mathbf{Y}^\mathbf{c}_{m,i})_{m=1}^{n_{\sigma}} \gets getData(i,n_q)$
        \State $X_i \gets (\mathbf{P}_{m,i},\mathbf{T}_{m,i},\mathbf{Y}^\mathbf{c}_{m,i})_{m=1}^{n_{\sigma}}$
        \State $(\mathbf{\Theta}_i,\mathbf{\Phi}_i,\mathbf{\Psi}_i) \gets getParams(X_i,n_q)$
        \State $E_i \gets getError((\mathbf{P}_{m,i},\mathbf{T}_{m,i})_{m=1}^{n_{\sigma}},\mathbf{\Theta}_i,\mathbf{\Phi}_i)$
        \If{$E_i<E^*_i$}
            \State $n_{q,i}^* \gets n_{q,i}$
            \State $E^*_i \gets E_i$
            \State $(\mathbf{\Theta}_i^*,\mathbf{\Phi}_i^*,\mathbf{\Psi}_i^*) \gets (\mathbf{\Theta}_{i},\mathbf{\Phi}_{i},\mathbf{\Psi}_{i})$
        \EndIf
    \EndFor
\EndFor
\end{algorithm}
\end{algorithmic}

\begin{figure}[t]
    \centering
    \includegraphics[width=1\columnwidth]{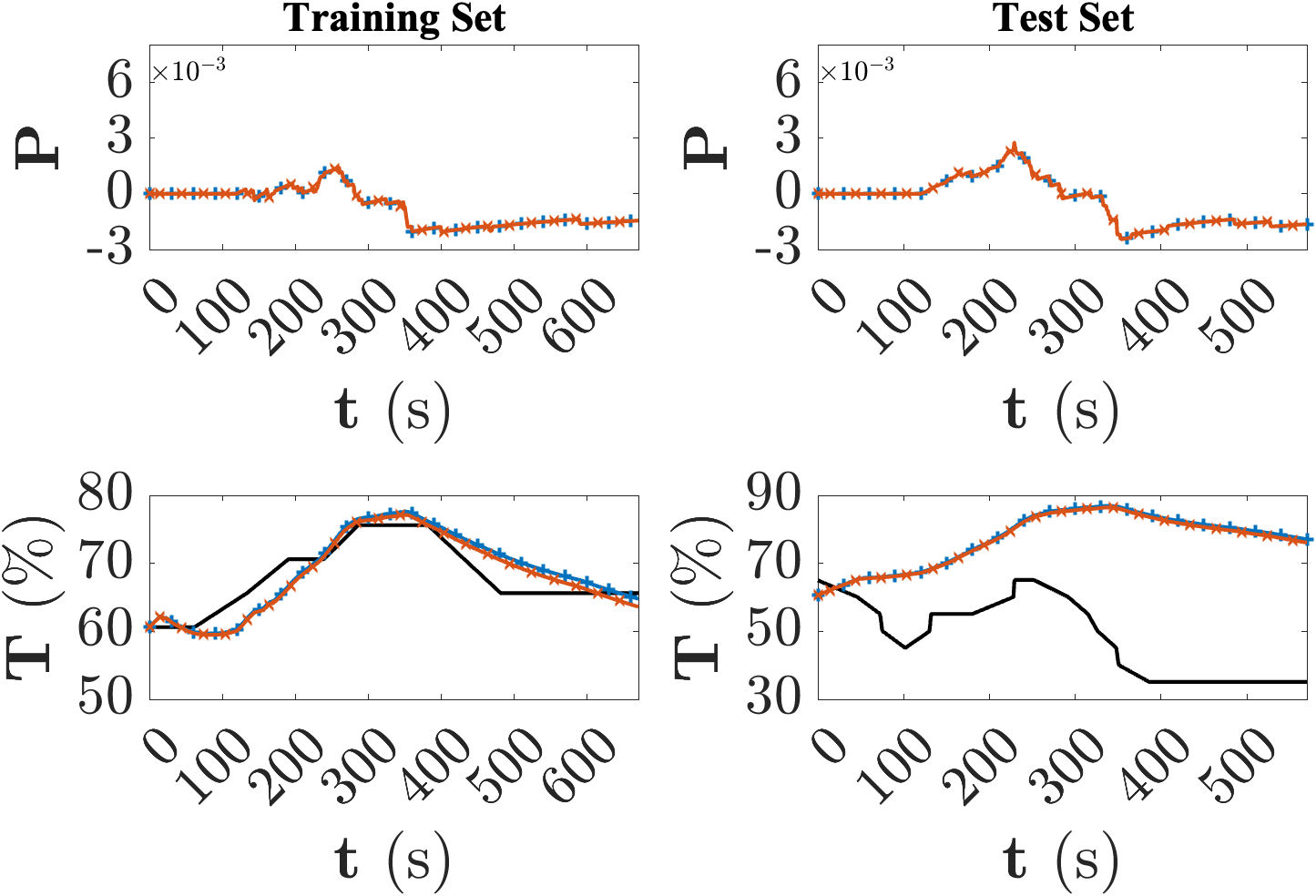}
    \caption{Autonomous system performance and predicted supervisor trust for participant 23; \textbf{—} ground truth, \textcolor{airforceblue}{\st{ + }} first-order model ($n_q^*=120.0$ s), \textcolor{red}{\st{ x }} second-order model ($n_q^*=120.0$ s).}
    \label{fig:individualTrustPredictions}
\end{figure}

\begin{figure}[t!]
    \centering
    \includegraphics[width=\columnwidth]{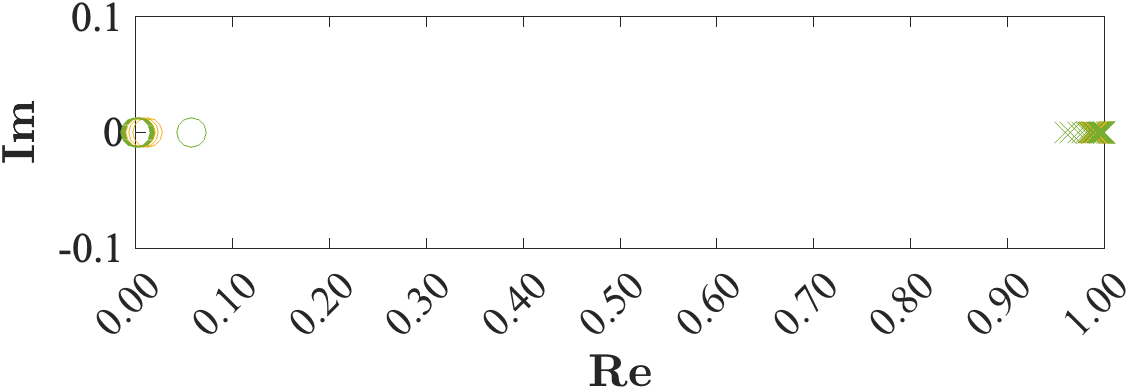}\\
    \caption{The distribution of participants' second-order $\mathbf{P}$-to-$\mathbf{T}$ transfer function poles (yellow: mode 2, green: mode 5; $\times$: pole 1, $\circ$: pole 2).}
    \label{fig:indpolsep1}
\end{figure}

\paragraph*{Individual Trust Models}

By setting $n_c=n_s$, defining mean squared error as the error metric, and using Algorithm \ref{alg:findnq} to solve \eqref{eq:optimAlpha}--\eqref{eq:optimC}, the optimal values for both $n_q^*$ and the parameters for the individual trust models were found.
Each individual's tuple $(n_{q,i}^*,\mathbf{\Theta}^*_i,\mathbf{\Phi}^*_i,\mathbf{\Psi}^*_i)$ was then used to generate predictions for the corresponding test set.

As an illustrative example, the individual trust model predictions for participant 23's training and test sets are displayed in Figure \ref{fig:individualTrustPredictions} beneath the corresponding performance signals. 
Notably the predictions of the second-order model are not greatly different to those of the first-order model for participant 23.
This observation holds for every participant surveyed, suggesting that the first-order model may capture sufficient information about the system dynamics.

An explanation for this observation follows from examining Figure \ref{fig:indpolsep1}.
There is a clear separation between two sets of poles: the weaker poles have magnitudes less than $0.10$, while the dominant poles have magnitudes between $0.9$ and $1.00$.
Consequently, a single-pole model is a reasonable approximation of the trust dynamics at the time scale of interest. 
For this reason, we only identify first-order models in this paper for the population and cluster-based models of trust dynamics.
\paragraph*{Trust Modeling for a Population}

\begin{figure}
    \centering
    \includegraphics[width=1\columnwidth]{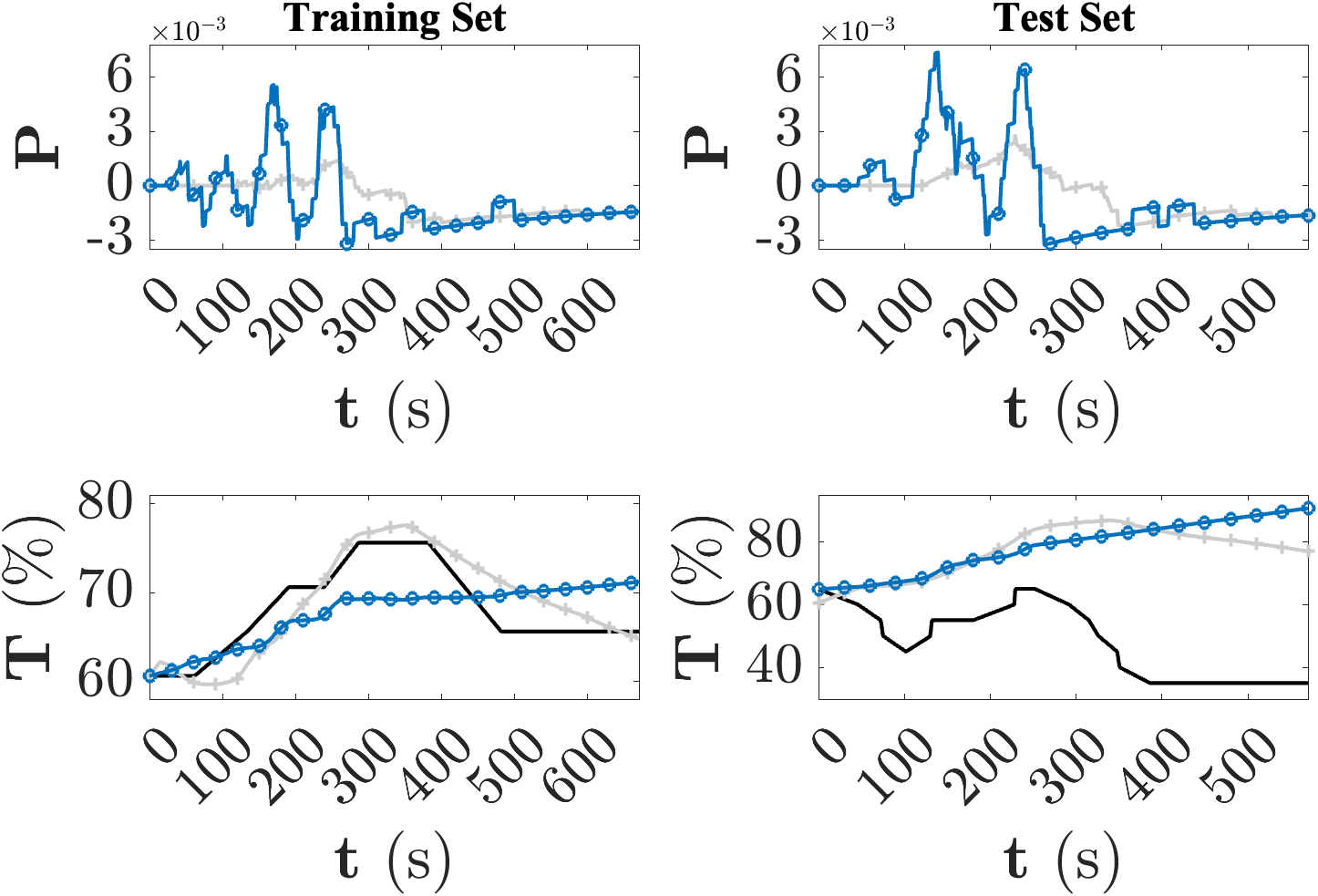}
    \caption{Autonomous system performance and predicted supervisor trust for participant 23 using the population model (\textbf{—} ground truth, \textcolor{airforceblue}{\st{ o }} first-order model with $n_q^*=30$ s) versus the individual model's predictions (\textcolor{navygray}{\st{ + }} first-order model with $n_q^*=120$ s).}
    \label{fig:populationTrustPredictions}
\end{figure}

\begin{figure}
    \centering
    \includegraphics[width=1\columnwidth]{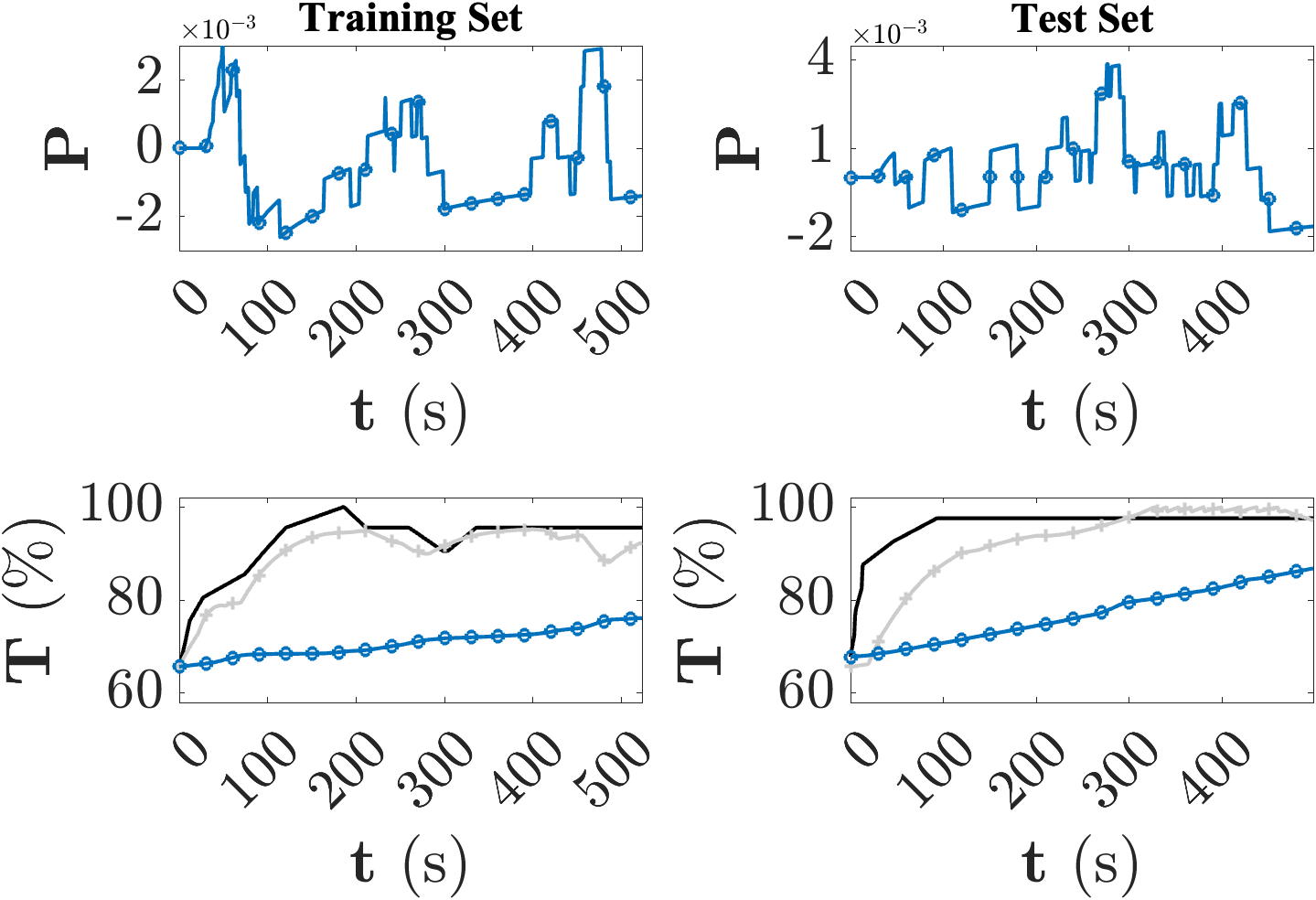}
    \caption{Autonomous system performance and predicted supervisor trust for participant 3 using the population model (\textbf{—} ground truth, \textcolor{airforceblue}{\st{ o }} first-order model with $n_q^*=30$ s) versus the individual model's predictions (\textcolor{navygray}{\st{ + }} first-order model with $n_q^*=30$ s).}
    \label{fig:populationTrustPredictions3}
\end{figure}

\begin{table}[t]
\centering
{\renewcommand{\arraystretch}{1.2}
\resizebox{1\columnwidth}{!}{
\begin{tabular}{|c|c|c|c|c|}
\hline
$m$ & $\bar{A}_m$ & $\bar{B}_m$ & $\bar{G}_m$ \\ \hline
1	& $9.90\times10^{-1}$	& $1.36\times10^{1}(1-\frac{\mathbf{T}-0.999}{0.001})$	& $0.00$	\\ \hline
2	& $1.00$	& $1.36\times10^{1}$	& $2.32\times10^{-2}$	\\ \hline
3	& $1.01$	& $1.36\times10^{1}(1-\frac{\mathbf{T}-0.999}{0.001})$	& $0.00$	\\ \hline
4	& $9.90\times10^{-1}$	& $1.11\times10^{1}(1-\frac{0.001-\mathbf{T}}{0.001})$	& $0.00$	\\ \hline
5	& $1.00$	& $1.11\times10^{1}$	& $2.56\times10^{-2}$	\\ \hline
6	& $1.01$	& $1.11\times10^{1}(1-\frac{0.001-\mathbf{T}}{0.001})$	& $0.00$	\\ \hline
\end{tabular}}}
\caption{Optimal population model parameters when $n_q^*=30$ s.}\label{tab:pop}
\end{table}
\begin{figure}[t]
    \centering
    \includegraphics[width=\columnwidth]{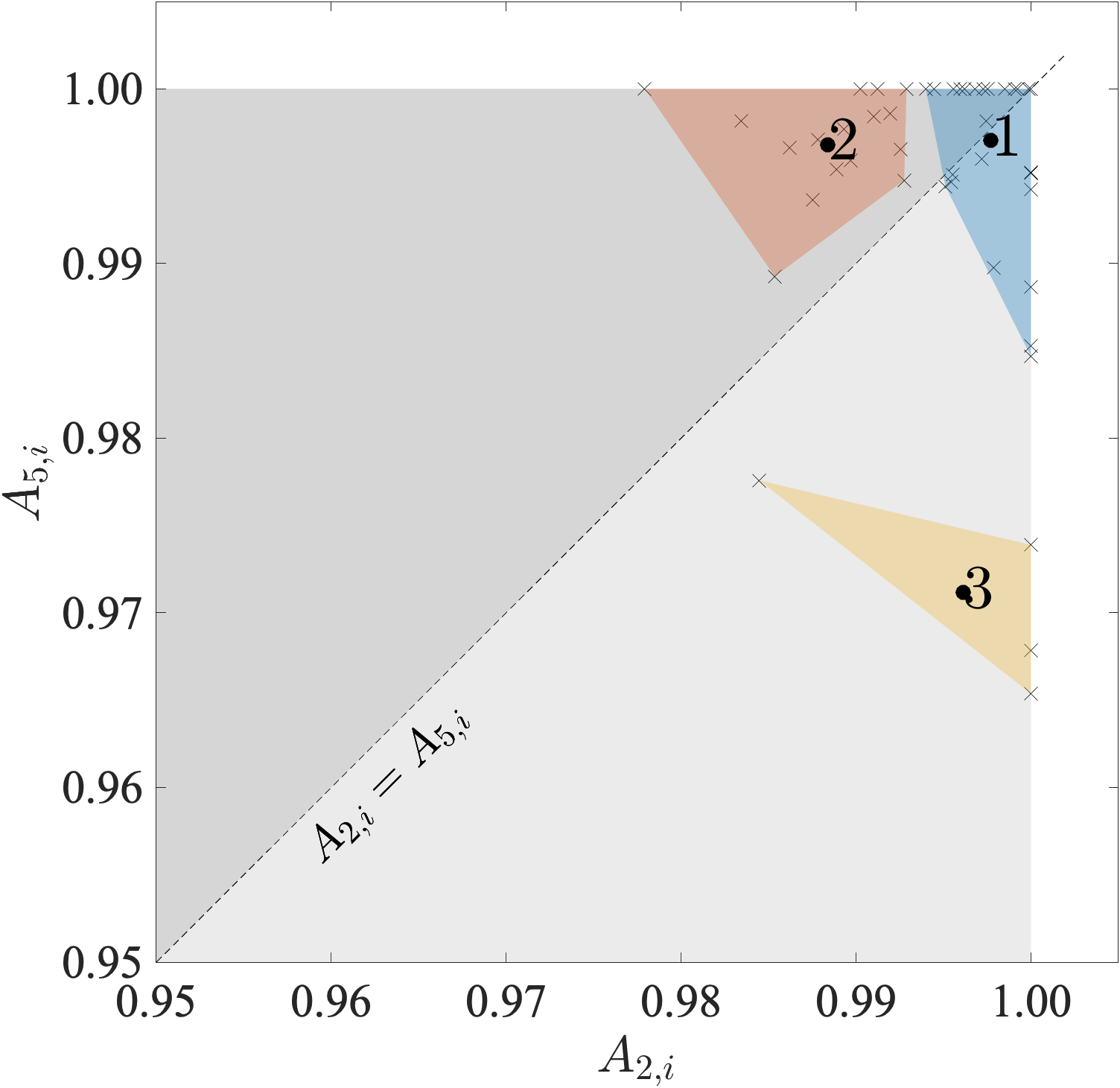}
    \caption{The first-order model poles for modes 2 and 5 are contained within $k=3$ clusters ($\times$ individual model datum, $\bullet$ \begin{hlb}weighted mean of cluster members' model parameters\end{hlb}).}
    \label{fig:clusters2D}
\end{figure}

Parameters for a first-order population trust model were found 
by setting $n_c=1$ (thus concatenating all the participants' data) and using Algorithm \ref{alg:findnq} with the mean squared error metric to solve \eqref{eq:optimAlpha}--\eqref{eq:optimC}.
The identified population model parameters appear in Table \ref{tab:pop}.

Comparing the predictions of the population model with those of the individual model for participant 23 in Figure \ref{fig:populationTrustPredictions}, we first observe that decreasing $n_q^*$ from $120$ s to $30$ s results in $\mathbf{P}$ varying faster and with greater amplitude.
As $\mathbf{P}$ is the input to the trust dynamics, the change in the memory length can result in very different model parameter values $(\bar{A}_m,\bar{B}_m)$ (contrast $A_{2,23} = 9.96\times10^{-1}$ and $B_{2,23} = 6.62\times 10^1$ with $\bar{A}_2 = 1.00$ and $\bar{B}_2 = 1.36\times10^{1}$), with consequences for the models' respective trust response trajectories.
Regarding hyperparameters, for all three types of models we have provided values for the memory length $n_q^*$ that have minimized the models' mean squared errors for training set predictions.
As demonstrated in Figure \ref{fig:populationTrustPredictions}, smaller values of $n_q$ yield $\mathbf{P}$ signals that spike faster and with larger amplitudes.
For a given supervisor this can influence the individual model's parameter values (and therefore dynamics), however it is $\mathbf{P}$ itself that has a dominant effect on parameter identification and trust inference. 

It is worth noting that changes in memory length are unlikely to be the sole nor most important reason for differences in parameter values between the population model and an individual's model.
Consider the trust responses of participant 3 depicted in Figure \ref{fig:populationTrustPredictions3}: both models use $n_q^*=30.0$ s and have the same $\mathbf{P}$ signal, however the two models' trust dynamics have different model parameter values and trajectories.
It appears more likely that the choice of training data used to identify the model parameters has a stronger influence than $n_q^*$ on the identified parameter values, resulting in the reduced accuracy of the population model predictions.

\paragraph*{Clustered Trust Models}

\begin{figure}[t]
    \centering
    \includegraphics[width=1\columnwidth]{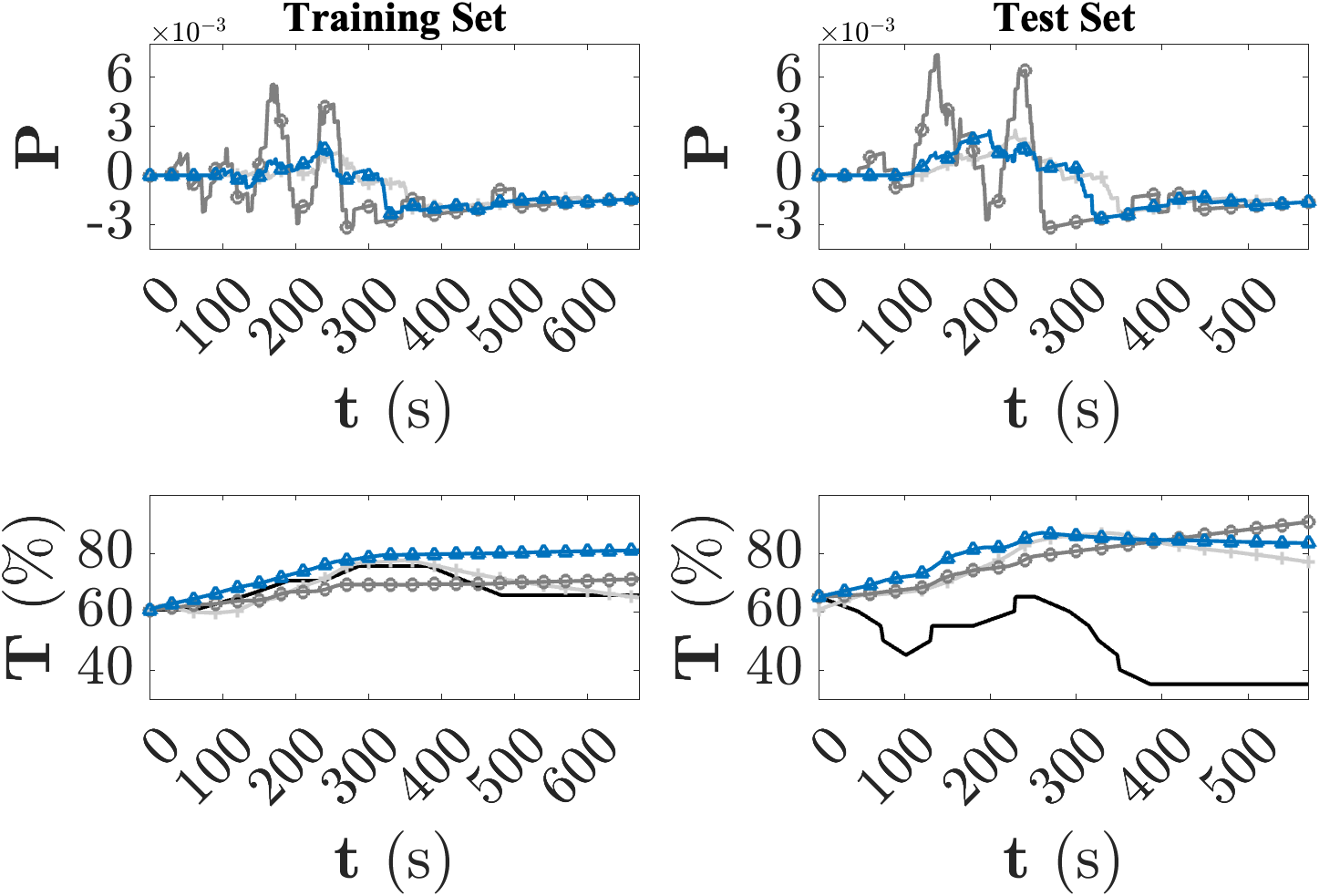}
    \caption{Autonomous system performance and predicted supervisor trust for participant 23 using the \textit{ambivalent} cluster model (\textbf{—} ground truth, \textcolor{airforceblue}{\st{ $\triangle$ }}: first-order model with $n_q^*=90.0$ s) versus the individual model's predictions (\textcolor{navygray}{\st{ + }} first-order model with $n_q^*=120$ s) and the population model's predictions (\textcolor{newgray}{\st{ o }} first-order model with $n_q^*=30$ s).}
    \label{fig:clusterTrustPredictions23}
\end{figure}
\begin{figure}[t]
    \centering
    \includegraphics[width=1\columnwidth]{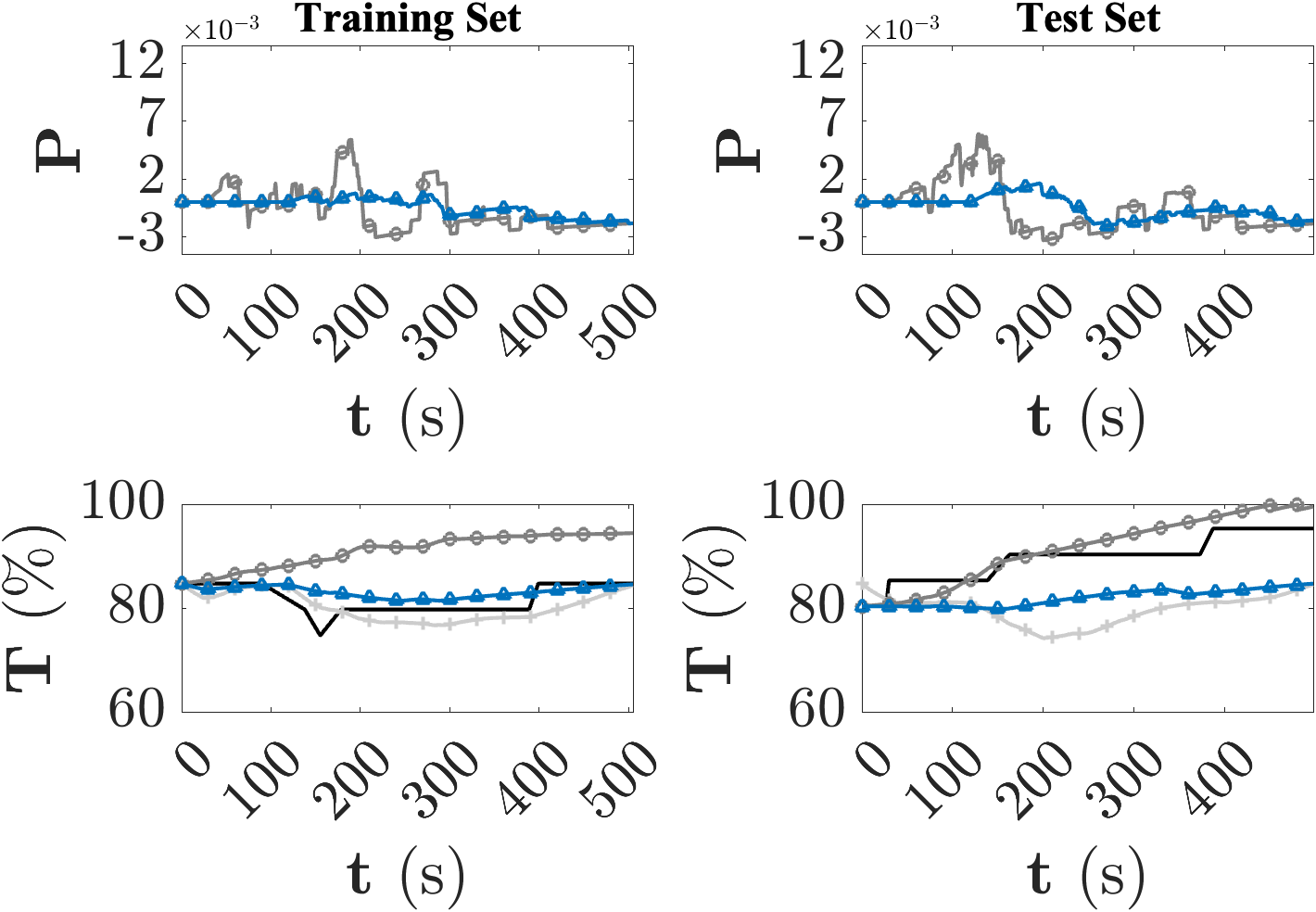}
    \caption{Autonomous system performance and predicted supervisor trust for participant 21 using the \textit{pessimistic} cluster model (\textbf{—} ground truth, \textcolor{airforceblue}{\st{ $\triangle$ }}: first-order model with $n_q^*=120.0$ s) versus the individual model's predictions (\textcolor{navygray}{\st{ + }} first-order model with $n_q^*=120$ s) and the population model's predictions (\textcolor{newgray}{\st{ o }} first-order model with $n_q^*=30$ s).}
    \label{fig:clusterTrustPredictions21}
\end{figure}
\begin{figure}[t]
    \centering
    \includegraphics[width=1\columnwidth]{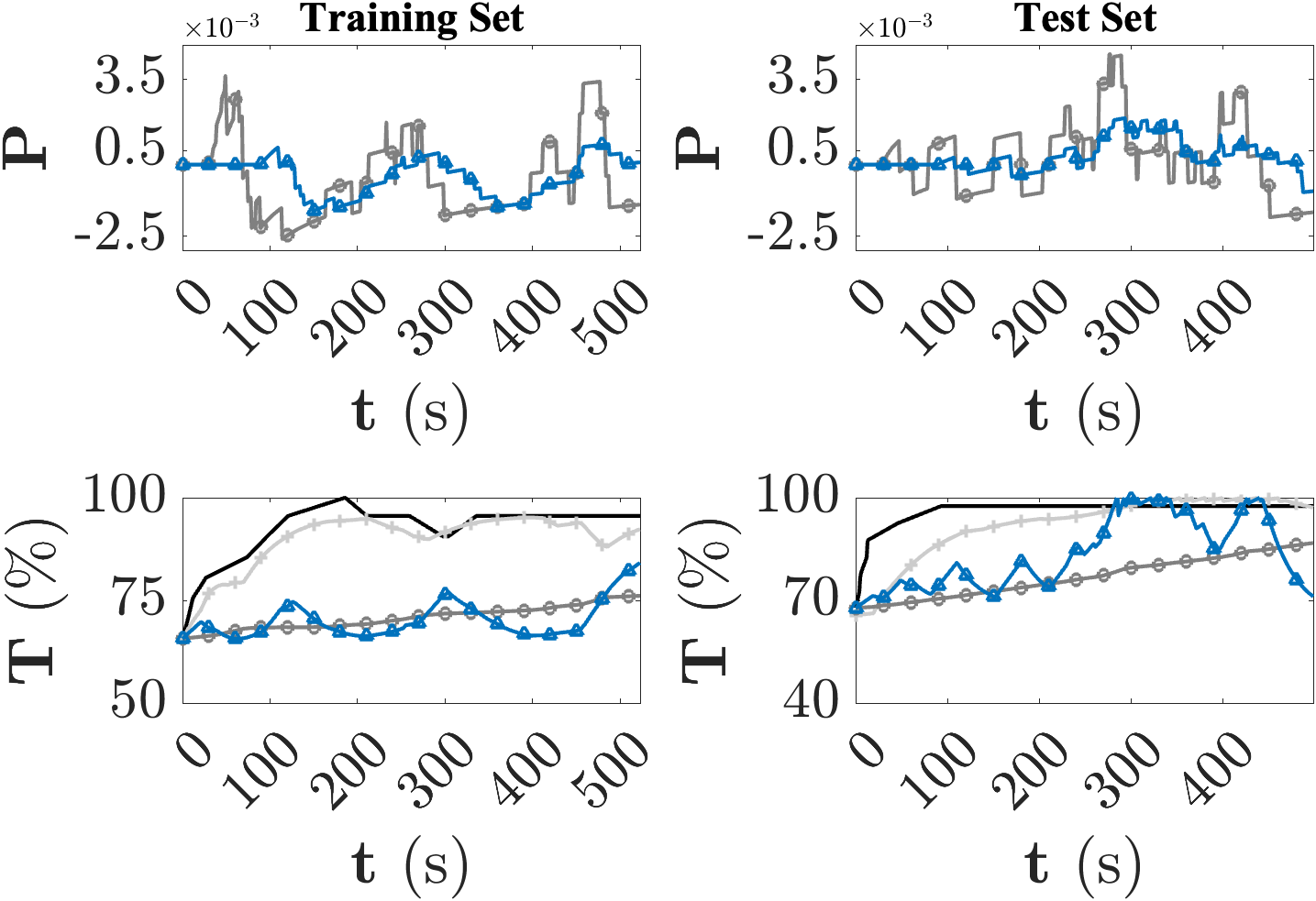}
    \caption{Autonomous system performance and predicted supervisor trust for participant 3 using the \textit{optimistic} cluster model (\textbf{—} ground truth, \textcolor{airforceblue}{\st{ $\triangle$ }}: first-order model with $n_q^*=90.0$ s) versus the individual model's predictions (\textcolor{navygray}{\st{ + }} first-order model with $n_q^*=30$ s) and the population model's predictions (\textcolor{newgray}{\st{ o }} first-order model with $n_q^*=30$ s).}
    \label{fig:clusterTrustPredictions3}
\end{figure}

\begin{table}[t]
\centering
{\renewcommand{\arraystretch}{1.2}
\resizebox{1\columnwidth}{!}{
\begin{tabular}{|c|ccc|}
\hline
\multicolumn{4}{|c|}{Cluster 1 (\textit{Ambivalent}, $n_q^* = 90.0$ s)} \\ \hline
\multicolumn{1}{|c|}{$m$} & \multicolumn{1}{c|}{$A_m^1$} & \multicolumn{1}{c|}{$B_m^1$} & \multicolumn{1}{c|}{$G_m^1$}\\ \hline
1& \multicolumn{1}{c|}{$9.90\times10^{-1}$}	& \multicolumn{1}{c|}{$2.42\times10^{1}(1-\frac{\mathbf{T}-0.999}{0.001})$}	& \multicolumn{1}{c|}{$0.00$}	\\ \hline 
2& \multicolumn{1}{c|}{$9.98\times10^{-1}$}	& \multicolumn{1}{c|}{$2.42\times10^{1}$}	& \multicolumn{1}{c|}{$2.20\times10^{-1}$}	\\ \hline 
3& \multicolumn{1}{c|}{$1.01$}	& \multicolumn{1}{c|}{$2.42\times10^{1}(1-\frac{\mathbf{T}-0.999}{0.001})$}	& \multicolumn{1}{c|}{$0.00$}	\\ \hline 
4& \multicolumn{1}{c|}{$9.90\times10^{-1}$}	& \multicolumn{1}{c|}{$8.88(1-\frac{0.001-\mathbf{T}}{0.001})$}	& \multicolumn{1}{c|}{$0.00$}	\\ \hline 
5& \multicolumn{1}{c|}{$9.97\times10^{-1}$}	& \multicolumn{1}{c|}{$8.88$}	& \multicolumn{1}{c|}{$2.58\times10^{-1}$}	\\ \hline 
6& \multicolumn{1}{c|}{$1.01$}	& \multicolumn{1}{c|}{$8.88(1-\frac{0.001-\mathbf{T}}{0.001})$}	& \multicolumn{1}{c|}{$0.00$}	\\ \hline 
\multicolumn{4}{|c|}{Cluster 2 (\textit{Pessimistic}, $n_q^* = 120.0$ s)} \\ \hline
\multicolumn{1}{|c|}{$m$} & \multicolumn{1}{c|}{$A_m^2$} & \multicolumn{1}{c|}{$B_m^2$} & \multicolumn{1}{c|}{$G_m^2$}\\ \hline
1& \multicolumn{1}{c|}{$9.90\times10^{-1}$}	& \multicolumn{1}{c|}{$-2.72(1-\frac{\mathbf{T}-0.999}{0.001})$}	& \multicolumn{1}{c|}{$0.00$}	\\ \hline 
2& \multicolumn{1}{c|}{$9.88\times10^{-1}$}	& \multicolumn{1}{c|}{$-2.72$}	& \multicolumn{1}{c|}{$9.32\times10^{-1}$}	\\ \hline 
3& \multicolumn{1}{c|}{$1.01$}	& \multicolumn{1}{c|}{$-2.72(1-\frac{\mathbf{T}-0.999}{0.001})$}	& \multicolumn{1}{c|}{$0.00$}	\\ \hline 
4& \multicolumn{1}{c|}{$9.90\times10^{-1}$}	& \multicolumn{1}{c|}{$-4.78(1-\frac{0.001-\mathbf{T}}{0.001})$}	& \multicolumn{1}{c|}{$0.00$}	\\ \hline 
5& \multicolumn{1}{c|}{$9.97\times10^{-1}$}	& \multicolumn{1}{c|}{$-4.78$}	& \multicolumn{1}{c|}{$2.75\times10^{-1}$}	\\ \hline 
6& \multicolumn{1}{c|}{$1.01$}	& \multicolumn{1}{c|}{$-4.78(1-\frac{0.001-\mathbf{T}}{0.001})$}	& \multicolumn{1}{c|}{$0.00$}	\\ \hline 
\multicolumn{4}{|c|}{Cluster 3 (\textit{Optimistic}, $n_q^* = 90.0$ s)} \\ \hline
\multicolumn{1}{|c|}{$m$} & \multicolumn{1}{c|}{$A_m^3$} & \multicolumn{1}{c|}{$B_m^3$} & \multicolumn{1}{c|}{$G_m^3$}\\ \hline
1& \multicolumn{1}{c|}{$9.90\times10^{-1}$}	& \multicolumn{1}{c|}{$2.02\times10^{2}(1-\frac{\mathbf{T}-0.999}{0.001})$}	& \multicolumn{1}{c|}{$0.00$}	\\ \hline 
2& \multicolumn{1}{c|}{$9.96\times10^{-1}$}	& \multicolumn{1}{c|}{$2.02\times10^{2}$}	& \multicolumn{1}{c|}{$4.56\times10^{-1}$}	\\ \hline 
3& \multicolumn{1}{c|}{$1.01$}	& \multicolumn{1}{c|}{$2.02\times10^{2}(1-\frac{\mathbf{T}-0.999}{0.001})$}	& \multicolumn{1}{c|}{$0.00$}	\\ \hline 
4& \multicolumn{1}{c|}{$9.90\times10^{-1}$}	& \multicolumn{1}{c|}{$1.14\times10^{2}(1-\frac{0.001-\mathbf{T}}{0.001})$}	& \multicolumn{1}{c|}{$0.00$}	\\ \hline 
5& \multicolumn{1}{c|}{$9.71\times10^{-1}$}	& \multicolumn{1}{c|}{$1.14\times10^{2}$}	& \multicolumn{1}{c|}{$2.04$}	\\ \hline 
6& \multicolumn{1}{c|}{$1.01$}	& \multicolumn{1}{c|}{$1.14\times10^{2}(1-\frac{0.001-\mathbf{T}}{0.001})$}	& \multicolumn{1}{c|}{$0.00$}	\\ \hline  
\end{tabular}}}
\caption{\begin{hlb}Cluster centroid parameters, after clustering performed in the $(A_2, A_5)$-plane.\end{hlb}}\label{tab:clus}
\end{table}

In order to apply Algorithm \ref{alg:findnq} the number of groups $n_c$ must be specified, however the optimal value of $n_c$ for a population (i.e. the number of clusters) may be unknown \textit{a priori}.
To determine the optimal number of clusters, $k$-means clustering was performed for $k\in\{2,...,10\}$ using 1000 replicates and the $k$-means$++$ cluster initialization algorithm \cite{arthur_kmeanspp_2006}.
For each value of $k$, the replicate yielding the lowest sum of distances between each individual's datum and the nearest cluster was selected.
For $k>3$, clusters emerge that consist of single individuals, which is undesirable when seeking to identify the main sub-groups in populations. 
For this reason $k=3$ was chosen to ensure that the cluster models contained at least two individuals.
With significantly more participants, it may be possible to identify and exclude individuals exhibiting idiosyncratic behaviors during cluster determination.

The three clusters (which we qualitatively characterize as \textit{ambivalent}, \textit{pessimistic}, and \textit{optimistic} with respect to the autonomous system's performance) can be clearly observed in Figure \ref{fig:clusters2D}.
Here the individual model parameters are plotted and clustered in the $(A_{2},A_{5})$ plane, with the cluster model parameters generated using the weighted mean of cluster members' model parameters.
The dashed line of equality indicates the locus of pole values for which $A_{2,i}=A_{5,i}$, denoting a degenerate case that accommodates the unswitched linear models proposed in prior literature.
The dark gray region above the dashed line denotes a `quick to lose, slow to gain' trust response, while the light gray region below denotes a `quick to gain, slow to lose' trust response.
A summary of the parameter values identified for each cluster is displayed in Table \ref{tab:clus}.

The \textit{ambivalent} cluster straddles the dashed line, containing the population model and $59.2\%$ of participants. 
This cluster's trust responses are typified by those of the population model responses as demonstrated in Figure \ref{fig:clusterTrustPredictions23}.
Clustering individual trust responses according to the values of $A_{m,i}$ (as depicted in Figure \ref{fig:clusters2D}) would give the impression of a symmetric trust response.
However, in comparing the values of $B_{m,1}$ we observe appreciable differences between modes 2 and 5.
This implies that there may be an advantage to incorporating an asymmetrical response to $\mathbf{P}$.
To accommodate this asymmetry, a switched-linear model is therefore necessary.

The \textit{pessimistic} cluster lies entirely within the dark gray region with $32.6\%$ of individuals belonging to this cluster.
The centroid of this cluster is further away from the population model than that of the \textit{ambivalent} cluster, hence for these participants the cluster model predictions should be closer to the individual's measured trust response than the population model predictions (as exemplified for participant 21 in Figure \ref{fig:clusterTrustPredictions21}).

The \textit{optimistic} cluster located in the light gray region contains $8.1\%$ of participants.
Similar to the \textit{pessimistic} cluster, we observe in Figure \ref{fig:clusterTrustPredictions3} for participant 3 that the cluster model predictions are closer to the individual's measured trust response than those of the population model.

\begin{figure}[t]
    \centering
    \includegraphics[width=\columnwidth]{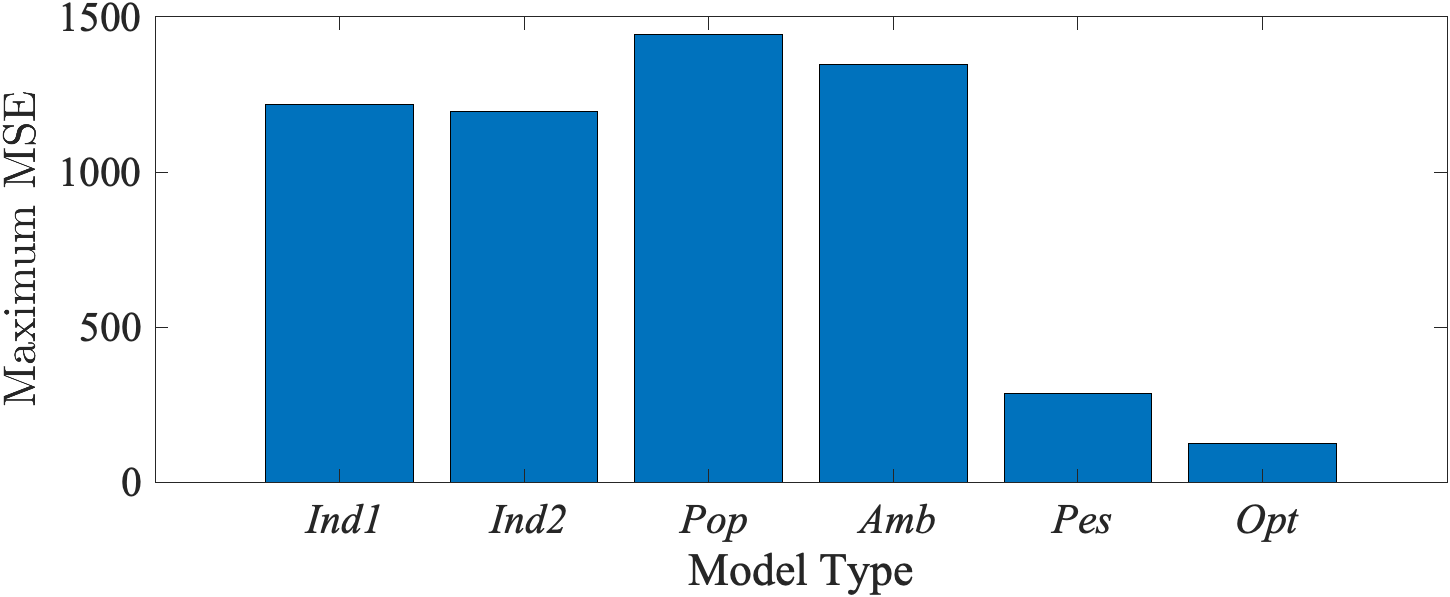}
    \caption{Maximum mean squared error statistics for test set predictions.\label{fig:mse_max}}
\end{figure}

To provide a broader comparison of the three model types in this paper (individual, population, and cluster-based), we consider the mean squared error of a model's predictions for unseen data in the test sets.
Denote the $i$th supervisor's measured test set trust response $\mathbf{T}_i[k]$, $k\in\{1,...,n_k\}$.
Using the identified parameters found in Tables \ref{tab:ind}, \ref{tab:pop}, and \ref{tab:clus} with the individual, population and cluster models respectively, we generate the predicted test set trust responses $\hat{\mathbf{T}}_i[k]$ using \eqref{eq:trustupdate_sls} and \eqref{eq:switcher} with $n_T=1$.
The model's mean squared error is 
\begin{align}
    MSE(i) = \frac{1}{n_k}\sum_{k=1}^{n_k}(\mathbf{T}_i[k] - \hat{\mathbf{T}}_i[k])^2.
\end{align}
The resulting maximum values of $MSE(i)$ for the first- and second-order individual models \textit{Ind1} and \textit{Ind2}, population model \textit{Pop}, and cluster models \textit{Amb}, \textit{Pes}, and \textit{Opt} are depicted in Figure \ref{fig:mse_max}.

As might be reasonably expected, we observe that \textit{Pop} yields a larger maximum MSE than \textit{Ind1} and \textit{Ind2}, indicating the benefit of personalizing a trust model for a given supervisor. 

We also note that \textit{Amb}, \textit{Pes}, and \textit{Pop} all yield a lower maximum MSE than that of \textit{Pop}, demonstrating that the cluster models can better capture a variety of supervisor trust dynamics than a single population model, thereby justifying their use when cluster assignment is possible.

Perhaps unintuitively, \textit{Pes} and \textit{Opt} yield a lower maximum MSE than those of \textit{Ind1} and \textit{Ind2}.
We attribute this observation to individuals exhibiting differences in trust responses between the training and test sets, resulting in the personalized individual model overfitting the training set.
By clustering supervisors with similar trust responses before model identification, the volume and spread of training data used for identifying cluster model parameters is increased, in this case improving model robustness.
With a larger dataset containing more individuals, a potential extension could be to increase the number of clusters and exclude singletons, thus reducing outliers' influence on the main clusters' MSE.

Following these results we hypothesize that it would be advantageous to adopt a cluster-based model of trust if it is easier to classify a new individual than perform a complete model identification for the individual. 
In addition, from the results presented here it appears that the accuracy of trust prediction may improve compared to individual and population-wide model approaches subject to an appropriate cluster selection.
Such a switched linear trust model may subsequently be used in interfaces as depicted in Figure \ref{fig:framework}.

\section{Conclusions}\label{sec:conclusions}


In this paper we have sought to model a human supervisor's trust in an autonomous robotic system with event-triggered sampling of system status updates and supervisor interventions.
A novel application of a switched-linear system structure with dedicated saturation modes was proposed to represent the response of supervisor trust to autonomous system performance.
In a user study with 51 participants, parameters for individual supervisor trust models, a population trust model, and three clustered trust models were identified.

Simulations using the identified model parameters show that a first-order switched-linear model structure is appropriate for representing a variety of trust dynamics.
While for a majority of participants ($59.2\%$) the individual models reflected a trust dynamic described as \textit{ambivalent} with respect to performance, sizable minorities of participants exhibited \textit{pessimistic} trust dynamics ($32.6\%$) or \textit{optimistic} trust dynamics ($8.2\%$).
These results validate the existence of both symmetric and asymmetric trust responses described in prior literature.
In addition, the availability of population and cluster-based trust models now enables the real-time prediction of trust for unknown individuals.

In future work we will investigate an on-board observer of trust for the autonomous system informed by the three classes of models (individual, population, cluster) to estimate supervisor trust and adjust the autonomous system reference input.
It is expected that this closed feedback loop will allow the autonomous system to guide the human supervisor's trust towards an appropriate equilibrium.

\section*{Acknowledgements}
This research has been conducted on the lands of the Boonwurrung and Wurundjeri Woiwurrung people.

\bibliographystyle{IEEEtran}
\balance
\bibliography{main}
\end{document}